%% file: GPUSceneFlow.tex
\pdfoutput=1

\documentclass[10pt,twocolumn,letterpaper]{article}

\usepackage[utf8]{inputenc}
\usepackage[T1]{fontenc}
\usepackage{textcomp}

\usepackage{3dv}
\usepackage{times}
\usepackage{graphicx}
\usepackage{amsmath}
\usepackage{amssymb}

\usepackage{listings}
\usepackage{algorithm}
\usepackage{algorithmic}

\usepackage{microtype}
\usepackage{enumitem}
\input{macros}

\newcommand{\norm}[1]{\left\lVert#1\right\rVert}

\usepackage[pagebackref=true,breaklinks=true,letterpaper=true,colorlinks,bookmarks=false]{hyperref}

\usepackage[numbers,sort]{natbib}
\usepackage[capitalise]{cleveref}
\Crefname{figure}{Figure}{Figures}
\crefformat{equation}{Equation~#2#1#3}
\crefmultiformat{equation}{Equations~#2#1#3}{ and~#2#1#3}{, #2#1#3}{ and~#2#1#3}

\threedvfinalcopy 

\def\threedvPaperID{124} 

\begin{document}

\title{Real-time Halfway Domain Reconstruction of Motion and Geometry}

\author{
		Lucas Thies$^\text{1}$~~~~Michael Zollhöfer$^\text{2}$~~~~Christian Richardt$^\text{2,3,4}$~~~~Christian Theobalt$^\text{2}$~~~~Günther Greiner$^\text{1}$ \vspace{0.1cm}\\
		{\small $^\text{1}$University of Erlangen-Nuremberg~~~$^\text{2}$Max Planck Institute for Informatics~~~$^\text{3}$Intel Visual Computing Institute~~~$^\text{4}$University of Bath}\vspace{0.1cm}\\
	{\tt\small \{lucas.thies,guenther.greiner\}@fau.de}~~
	{\tt\small \{mzollhoef,richardt,theobalt\}@mpi-inf.mpg.de}
}

\hypersetup{
	pdftitle={Real-time Halfway Domain Reconstruction of Motion and Geometry},
	pdfauthor={Lucas Thies, Michael Zollhöfer, Christian Richardt, Christian Theobalt, Günther Greiner}
}


\maketitle

\input{0abstract}
\input{1introduction}
\input{2relatedwork}
\input{3representation}
\input{5energy}
\input{6optimization}
\input{7results}
\input{8conclusion}

\subsection*{Acknowledgements}

We thank Anna Hilsmann for the high-quality stereo pairs, and Levi Valgaerts for the high-quality stereo sequences.
This research is funded by the ERC Starting Grant 335545 CapReal.

\appendix
\section{Parameter Settings} \label{sec:appendix}

\begin{table}[h!]
	\centering
	\begin{tabular}{ | c | c | c | c |}
		\hline
		 & Figure 2 & Figures 5, 6, 8 & Figures 7, 9 \\ \hline
		 $w_\text{reg}$ & 1.0 & 0.5 & 5.0  \\ \hline
		 $w_\text{photo}$ & 1.0 & 0.5 & 1.0 \\ \hline
		 $w_\text{grad}$ & 2.0	& 5.0 & 5.0 \\ \hline
		 $w_\text{epi}$ & 0.0 & 0.5 & 0.5 \\ \hline
		 $w_\text{s}$ & 5.0 & 0.75 & 0.5 \\ \hline
		 $w_\text{m}$ & 5.0 & 0.5	& 1.0 \\ \hline
		 $w_\text{d}$ & 0.5 & 0.01	& 1.0 \\ \hline
		 $m_\text{s}$ & 5.0	& 0.5 & 0.1 \\ \hline
		 $m_\text{m}$ & 100.0	& 10.0 & 10000.0 \\ \hline
		 $m_\text{d}$ & 1000.0	& 100.0 & 10000.0 \\ \hline
	\end{tabular}
	\vspace{0.1cm}
	\caption{All used parameters.}
	\label{tab:parameters}
\end{table}

The specific choice of parameters influences our scene flow energy and the reconstruction results.
Our approach is quite robust to variation in the specific parameter values.
Nevertheless, the best reconstruction results are obtained at the sweet spot between the data fitting term and the prior constraints.
\Cref{tab:parameters} provides the parameter settings used to generate the results.
The two parameters $w_\text{smooth} \!=\! 1$ and $w_\text{mag} \!=\! 1$ are constant during all experiments.

{\small
\bibliographystyle{ieee}
\bibliography{GPUSceneFlow}
}

\end{document}

%% file: macros.tex
\definecolor{Brown}{cmyk}{0,0.81,1,0.60}
\definecolor{OliveGreen}{cmyk}{0.64,0,0.95,0.40}
\definecolor{CadetBlue}{cmyk}{0.62,0.57,0.23,0}
\definecolor{lightlightgray}{gray}{0.9}
\definecolor{gray}{gray}{0.4}
\definecolor{skyblue}{rgb}{0.2,0.6,0.9}
\definecolor{placeholder}{rgb}{0.6,0.8,0.95}

\newcommand{\IGNORE}[1]{{}}

\newcommand{\NEW}[1]{#1}

\newcommand{\inlineheading}[1]{\vspace{0.5em}\noindent\textbf{#1\hspace{0.5em}}}

\newcommand{\filluptopage}[1]{%
  \clearpage
  \loop\ifnum\value{page}<#1\relax
    \null\clearpage
  \repeat
  \loop\ifnum\value{page}=#1\relax
    \null\clearpage
  \repeat
}

\DeclareMathOperator*{\argmin}{argmin}

\def \path{\bp C}



%% file: 0abstract.tex

\begin{abstract}
We present a novel approach for real-time joint reconstruction of 3D scene motion and geometry from binocular stereo videos.
Our approach is based on a novel variational halfway-domain scene flow formulation, which allows us to obtain highly accurate spatiotemporal reconstructions of shape and motion.
We solve the underlying optimization problem at real-time frame rates using a novel data-parallel robust non-linear optimization strategy.
Fast convergence and large displacement flows are achieved by employing a novel hierarchy that stores delta flows between hierarchy levels.
High performance is obtained by the introduction of a coarser warp grid that decouples the number of unknowns from the input resolution of the images.
We demonstrate our approach in a live setup that is based on two commodity webcams, as well as on publicly available video data.
Our extensive experiments and evaluations show that our approach produces high-quality dense reconstructions of 3D geometry and scene flow at real-time frame rates, and compares favorably to the state of the art.
\end{abstract}

%% file: 1introduction.tex

\section{Introduction}
\vspace{-0.5em}

Many tasks in computer vision, such as performance capture, free-viewpoint video and 3D motion understanding, require dynamic scene reconstruction from only a few video cameras.
%
Dynamic scene reconstruction comprises the estimation of 3D geometry and its motion over time, which has been coined \emph{scene flow} by Vedula et al. \cite{VedulBRCK2005}, in analogy to `optical flow' which describes 2D motion of points over time.
The 3D motion of points cannot be accurately estimated in isolation from the 3D geometry as depth information 
is required for computing the 3D motion of points.
Unlike structure-from-motion, scene flow does not assume a static scene, but objects in the scene can move about freely and deform non-rigidly.
The estimation of scene flow from RGB images in a geometrically well-constrained way
therefore requires as input two sets of stereo (binocular) images for consecutive time steps.
In recent years, scene flow has been an important ingredient in many real-world applications, including those mentioned before, like 3D motion understanding in automotive scenarios \cite{WedelBVRFC2011,MenzeG2015}, facial performance capture \cite{ValgaWBST2012,WuSVT2013} 
and free-viewpoint video \citep{LipskKM2014}. 

Recently, real-time capable approaches for computing scene flow from specialized RGB-D cameras were proposed. 
However, existing approaches for computing dense scene flow from RGB images (without depth) require considerable computation time, in the order of minutes per frame (see KITTI scene flow evaluation 2015 \cite{MenzeG2015}).
This is because most dense scene flow approaches use variational formulations that result in large systems of equations with millions of unknowns that are computationally expensive to solve, despite efficient coarse-to-fine hierarchical optimization schemes.
The high computational complexity severely limits the applicability of these existing approaches.

In this paper, we thus propose the first approach for estimating dense scene flow and scene geometry at real-time rates ($\geqslant$30\,Hz) from binocular RGB video.
Even with the computational processing power of modern GPUs, existing dense binocular approaches are far from real-time performance, or achieve at best near-real-time rates using FPGAs \cite{WedelBVRFC2011}.
To achieve the real-time goal on a standard computer, we therefore introduce a new scene flow parameterization in terms of a spatiotemporal halfway domain that lies conceptually halfway between both camera viewpoints and between the two time steps (see flow illustration in \cref{fig:flow}).
In addition, we propose a novel, mesh-based coarse-to-fine warping scheme that accumulates pixel-level evidence within grid cells while dramatically reducing the number of unknown flow variables that need to be optimized.
During the coarse-to-fine warping, we leverage the GPU to efficiently bootstrap the computation of occlusions masks and illumination correction maps.
%
\NEW{We implemented} a new data-parallel optimization strategy that incorporates robust norms on a commodity graphics card\NEW{, which enables} our scene flow technique \NEW{to be} the first dense RGB-only method to achieve real-time frame rates (30\,Hz).
We show reconstruction results obtained with our live stereo webcam setup.
In addition, we compare to the scene flow approach of Valgaerts et al.~\cite{ValgaBZWST2010} (on the datasets of Valgaerts et al.~\cite{ValgaWBST2012}) and show reconstruction results on high-quality stereo pairs \cite{SchneKHE2011,BlumenthalBarby201489}.


%% file: 2relatedwork.tex

\section{Related Work} \label{sec:related}


\textbf{Stereo correspondence} – the computation of a disparity map from two rectified input images – is a long-standing problem in computer vision that has seen a large variety of techniques published over the last few decades \cite{BarnaF1982,ScharS2002}.
While our approach is not primarily aimed at computing just stereo disparity maps, as we also compute 3D scene flow over time, our approach can be adapted to stereo reconstruction by disabling the temporal component.
We therefore start by briefly reviewing the most relevant work from the stereo literature.
The first real-time approaches for stereo correspondence required custom hardware \cite{BertoB1998}, but the advent of graphics processing units (GPUs) made it even easier to achieve real-time performance.
At first, techniques implemented simple local stereo matching approaches with different cost aggregation schemes such as sum-of-squared-differences \cite{YangP2003}, adaptive aggregation \cite{WangLGYN2006} or others \cite{GongYWG2007}.
Later, more advanced stereo matching techniques were also ported and adjusted to work efficiently on GPUs, such as hierarchical belief propagation \cite{YangWYWLN2006} or adaptive support weights using the bilateral grid \cite{RichaODCD2010}.
However, these approaches often sacrifice quality for speed, which is an inherent trade-off.
High-resolution, high-quality disparity maps can for example be computed with approaches based on bilateral space stereo \cite{BarroASH2015} or mesh-based image warping \cite{SchneKHE2011,SchneKHE2011b,ZhangLCCCR2015,BlumenthalBarby201489}.
Most recently, deep neural networks have shown remarkable performance in stereo correspondence finding \cite{LuoSU2016,ZbontL2016,ChoyGSC2016}, and even directly estimating homographies \cite{DeTonMR2016}.




The goal of \textbf{scene flow} techniques is to compute the motion within a scene over time, for every visible 3D point between two time steps.
Many approaches have been proposed in recent years for computing scene flow from different visual input modalities, in particular RGB or RGB-D videos.
The proposed approaches include voxel coloring based on controlled multi-view camera setups \cite{VedulBRCK2005}, tracking of points and surfels \cite{DeverMG2006}, growing of correspondence seeds \cite{CechSH2011}, non-rigid scene registration \cite{BashaAHM2012}, particle-based estimation \cite{HadfiB2014}, semi-global \cite{YamagMU2014} \NEW{or wide-baseline matching \cite{RichaKVT2016}}.
The most common class of scene flow approaches, including ours, are variational methods, for both RGB \cite{HugueD2007,PonsKF2007,RabeMWF2010,ValgaBZWST2010,WedelBVRFC2011,BashaMK2013,HungXJ2013} and RGB-D inputs \cite{FerstRRB2014,SunSP2015,JaimeSSGC2015}, as they provide dense, continuous and strongly regularized solutions.
%

Many recent methods focus on estimating scene flow from RGB-D videos captured with consumer depth cameras \cite{LetouPB2011,HerbsRF2013,HadfiB2014,HornaFR2014,QuiroBDC2014,FerstRRB2014,SunSP2015,JaimeSSGC2015,ZanfiS2015}. Some of them also achieve real-time frame rates \cite{Jaimez2015,Bakkay:2015}, but in contrast to our RGB-based method, they use a special sensor to obtain depth maps.
%
%
The best-performing methods on the (RGB-only) KITTI 2015 scene flow benchmark \cite{MenzeG2015} enforce strong motion priors, like affine \cite{ZhangK2003} or piece-wise rigid motions \cite{VogelSR2015,MenzeG2015} that are ideal for the driving scenario.
However, our goal is to reconstruct general non-rigid dynamic scenes with a stereo camera pair, in which case many of these motion priors may be violated and counterproductive.
\NEW{Recently released datasets with synthetic ground truth also include non-rigid scenes \cite{MayerIHFCDB2016}.}
Like most previous binocular RGB approaches \cite[e.g.][]{ValgaBZWST2010,WedelBVRFC2011}, our technique computes scene flow between two consecutive time steps of stereo video.
As opposed to the RGB-D domain, dense binocular RGB-only variational scene flow computation at true real-time frame rates of 30\,Hz or more has not been shown so far.
Wedel et al. \cite{WedelBVRFC2011} achieved 20\,Hz for 320$\times$240 resolution videos using an implementation with a GPU and an FPGA. 
%


Scene flow estimation is also connected to non-rigid structure-from-motion \cite[e.g.][]{AvidaS2000,HartlV2008,ParkSMS2010,RusseYA2014,DaiLH2014,ZhengJDF2015}, 
although these approaches often apply strong motion priors and work best for small motions.
Another area of work related to ours is spatiotemporal stereo matching \cite{ZhangCS2003,RichaODCD2010,JiangLTZB2012}, which generally assumes static camera setups. 
As discussed in the introduction, scene flow is an essential ingredient for many applications, such as free-viewpoint video \cite{LipskKM2014}, 
facial performance capture \cite{ValgaWBST2012,WuSVT2013}, and 
motion understanding \cite{WedelBVRFC2011,MenzeG2015}.
Our work lifts the major computational barrier of previous scene flow approaches by demonstrating the first technique for real-time dense variational scene flow estimation from two RGB videos.


%% file: 3representation.tex

\section{Variational Halfway Domain Scene Flow}
\label{sec:parameterization}

Given two synchronized input camera streams (this can be achieved in hardware or software \citep[e.g.][]{MeyerSMP2008,HasleRTWGS2009,ElhaySKST2012,GaspaOF2014}), the goal of our dynamic scene reconstruction approach is to compute the dense 3D geometry and its motion over time.
%
%
In all our live experiments, we use a custom commodity stereo rig built using two Logitech HD Pro C920 webcams.
The captured live streams are assumed to be synchronized.

\begin{algorithm}[t]
\caption{Variational Halfway Domain Scene Flow}
\begin{algorithmic}
\STATE ($\mathcal{S}$, $\mathcal{O}$, $\mathcal{L}$) = Initialization();
\FOR{$i$ = 1 \ldots \ num\_levels}
\STATE $\mathcal{S}$ = Compute\_Scene\_Flow($\mathcal{S}$, $\mathcal{O}$, $\mathcal{L}$);
\STATE $\mathcal{O}$ = Occlusion\_Maps($\mathcal{S}$);
\STATE $\mathcal{L}$  = Illumination\_Maps($\mathcal{S}$);
\STATE Prolongation($\mathcal{S}$, $\mathcal{O}$, $\mathcal{L}$);
\ENDFOR
\end{algorithmic}
\label{alg:opt}
\end{algorithm}

Similar to Valgaerts et al. \cite{ValgaBZWST2010}, we parametrize scene flow using three unique flow fields: the stereo, motion and difference flow field.
We solve for the scene flow in a hierarchical coarse-to-fine fashion using a variational scene flow approach (see Algorithm \ref{alg:opt}).
During optimization, we bootstrap the computation of occlusion maps by rendering a triangulated version of the scene.
We also compute illumination correction maps based on the per-level results.
These occlusion and illumination correction maps are computed after the optimization on a level is finished and are upsampled (`prolongated') to the next finer hierarchy level to constrain the energy.
In the following, we provide more details on the used scene flow parameterization and how we check for flow validity.
Details on the illumination and occlusion map computation are provided in \cref{sec:optimization}.

\subsection{Halfway Domain Scene Flow Geometry}

\begin{figure}
  \centering
  \includegraphics[width=\linewidth]{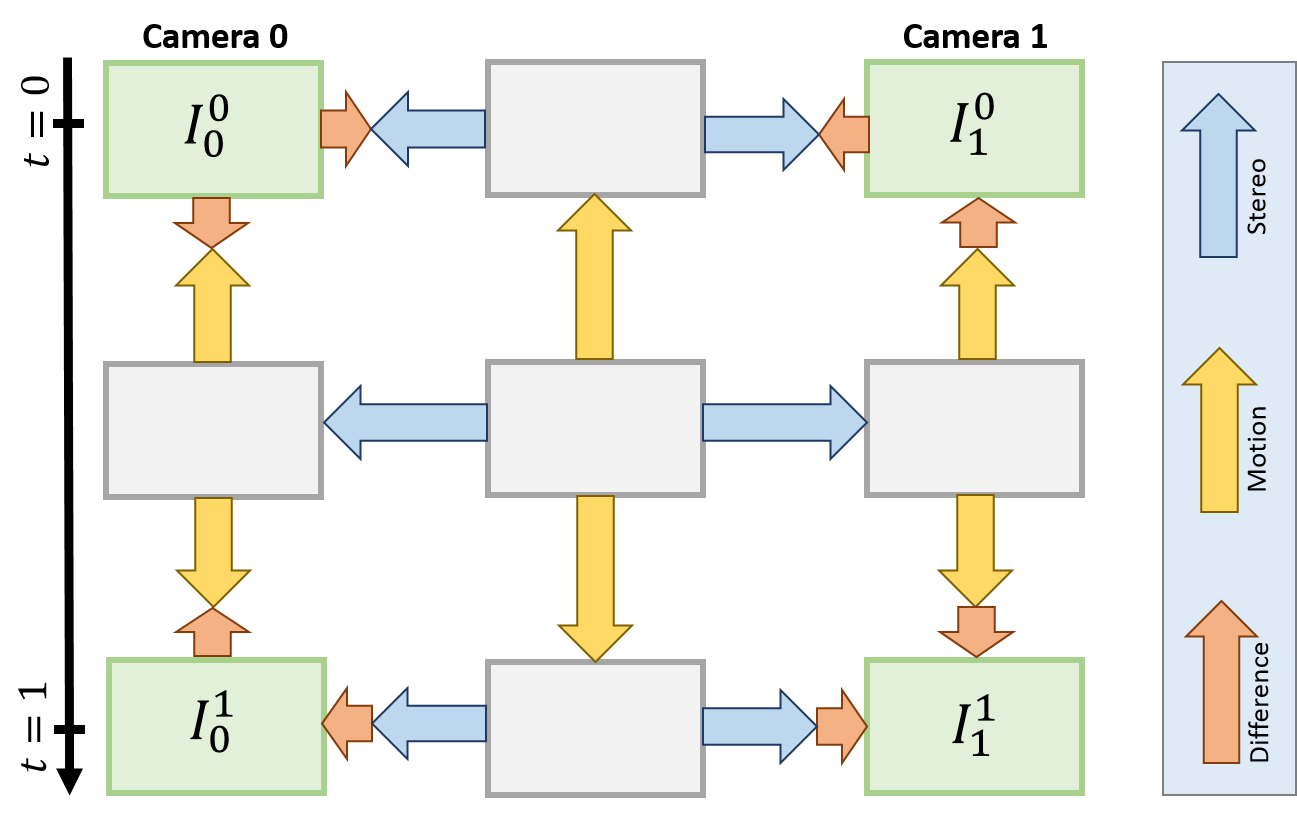}
  \caption{Binocular halfway-domain scene flow geometry.
  	Note that arrows pointing from left to right and top to bottom represent flows with positive signs, otherwise the sign is negative.}\vspace{-0.5em}
  \label{fig:flow}
\end{figure}

We extend the idea of the halfway correspondence domain \cite{LiaoLNHSY2014} to the context of scene flow, as illustrated in \cref{fig:flow}.
We consider two monochrome stereo image pairs $\mathcal{I}_c^{t}$, where $c \!\in\! \{0,1\}$ denotes the camera index (0: left, 1: right) and $t \!\in\! \{0,1\}$ denotes the time step (0: previous, 1: current).
The four captured images define the corners of the scene flow geometry.
The five in-between frames define intermediate states of warping between the captured images based on the flow data.
We define the scene flow $\mathcal{S}$ as a combination of three flows (stereo, motion and difference) relative to the halfway domain in the middle.
The two intermediate images to the left and right of the halfway domain can be thought of as being captured by virtual cameras at the halfway time step.
The top and bottom intermediate images can be thought of as being captured by a virtual in-between camera.
The pixels of the halfway domain (given by the integer pixel grid positions $\mathbf{x}_i \!\in\! \mathbb{N}^2$) can be mapped to the four input images by combining the per-pixel stereo $\{ \mathbf{s}_i \!\in\! \mathbb{R}^2\}_{i=1}^{N}$ (blue), motion $ \{ \mathbf{m}_i \!\in\! \mathbb{R}^2\}_{i=1}^{N} $ (yellow) and difference flow $ \{ \mathbf{d}_i \!\in\! \mathbb{R}^2\}_{i=1}^{N}$ (red), where $N$ is the number of pixels in the image.
The direction of the arrows indicates the target space of the flow field.
Arrows pointing from left to right and top to bottom represent positive signs, otherwise the sign is negative.

\subsection{Binocular Scene Flow Consistency}
\label{sec:consistency}

We consider the flows between all combinations of input frame pairs consistent if every pixel of one input image is mapped to the corresponding pixel in all other input images that see the same 3D surface point.
The consideration of all different mappings between the four input images gives rise to a total of six different consistency checks.
Note that we only model the checks in forward direction for higher efficiency.
Since checking for the same surface point is impossible, we at first relax the consistency condition to a brightness constancy check.
The first two checks are the stereo flow consistency checks, which map between the two cameras of the stereo pairs at \NEW{corresponding} time steps:
\begingroup\makeatletter\def\f@size{8}\check@mathfonts
\begin{align}
d_{0}(\mathbf{x}_i) &= \mathcal{I}_1^{0}(\mathbf{x}_i+\mathbf{s}_i- \mathbf{m}_i- \mathbf{d}_i)-\mathcal{I}_0^{0}(\mathbf{x}_i- \mathbf{s}_i- \mathbf{m}_i+\mathbf{d}_i) \text{,}\\
d_{1}(\mathbf{x}_i) &= \mathcal{I}_1^{1}(\mathbf{x}_i+\mathbf{s}_i+\mathbf{m}_i+\mathbf{d}_i)-\mathcal{I}_0^{1}(\mathbf{x}_i- \mathbf{s}_i+\mathbf{m}_i- \mathbf{d}_i) \text{.}
\end{align}
\endgroup
The second \NEW{pair} of checks model motion flow consistency, which considers images captured by the same camera at \NEW{consecutive} time steps:
\begingroup\makeatletter\def\f@size{8}\check@mathfonts
\begin{align}
d_{2}(\mathbf{x}_i) &= \mathcal{I}_0^{1}(\mathbf{x}_i- \mathbf{s}_i+\mathbf{m}_i- \mathbf{d}_i)-\mathcal{I}_0^{0}(\mathbf{x}_i- \mathbf{s}_i- \mathbf{m}_i+\mathbf{d}_i) \text{,}\\
d_{3}(\mathbf{x}_i) &= \mathcal{I}_1^{1}(\mathbf{x}_i+\mathbf{s}_i+\mathbf{m}_i+\mathbf{d}_i)-\mathcal{I}_1^{0}(\mathbf{x}_i+\mathbf{s}_i- \mathbf{m}_i- \mathbf{d}_i) \text{.}
\end{align}
\endgroup
Finally, the cross consistency checks consider the images captured by different cameras at different time steps:
\begingroup\makeatletter\def\f@size{8}\check@mathfonts
\begin{align}
d_{4}(\mathbf{x}_i) &= \mathcal{I}_1^{1}(\mathbf{x}_i+\mathbf{s}_i+\mathbf{m}_i+\mathbf{d}_i)-\mathcal{I}_0^{0}(\mathbf{x}_i- \mathbf{s}_i- \mathbf{m}_i+\mathbf{d}_i) \text{,}\\
d_{5}(\mathbf{x}_i) &= \mathcal{I}_0^{1}(\mathbf{x}_i- \mathbf{s}_i+\mathbf{m}_i- \mathbf{d}_i)-\mathcal{I}_1^{0}(\mathbf{x}_i+\mathbf{s}_i- \mathbf{m}_i- \mathbf{d}_i) \text{.}
\end{align}
\endgroup
Here we formulated consistency only in terms of brightness constancy.
In the following, we will also consider gradient constancy to define a matching criterion that is more robust to appearance and lighting changes.

\subsection{Scene Flow Parameterization}

%
Different from previous approaches, we parameterize the per-pixel flow fields based on a uniform deformation lattice 
with a coarser resolution than the captured input images, to achieve real-time performance.
This helps to resolve ambiguities in flow computation, since multiple input pixel observations influence the unknown displacement at each grid point.
In addition, the introduction of this deformation proxy reduces the number of unknowns and thus leads to higher efficiency.
In all our experiments, we used a coarsening factor of 2, meaning that we have a deformation grid point on every second pixel.
Since we parameterize the per-pixel displacements based on a coarser resolution warp grid, we obtain in-between flow values via bilinear interpolation.
For example, the per-pixel stereo flow can be obtained based on the $G$ stereo flow deformation nodes $\{ \mathbf{g}^s_k \}_{k=1}^{G}$ by:
\begin{equation}	
\mathbf{s}_i = \sum_{k=1}^{G}{ \alpha^{s}_{i,k} \cdot \mathbf{g}^s_k} \text{.}
\end{equation}
Here, the $\alpha^s_{i,k}$ are the bilinear interpolation weights for the per-pixel stereo flow $\mathbf{s}_i$.
Note that for a particular $i$, $\alpha^s_{i,k}$ defines a sparse partition of unity over the $G$ grid points (only four $k$'s have non-zero $\alpha^s_{i,k}$ for any $i$).
Similar relations also hold for the motion and difference flow fields.

%% file: 5energy.tex

\section{Spatiotemporal Scene Flow Objective} \label{sec:energy}

Similar to previous work \cite{ValgaBZWST2010}, we cast finding the scene flow that best explains the binocular input images in two successive time steps as a variational energy minimization problem.
The objective function takes into account both spatial alignment of the inputs warped to the halfway-domain reference frame and the validity of the flow fields.
Therefore, our non-linear scene flow objective $E$ is a mixture of spatial alignment $E_\text{align}$ and regularization constraints $E_\text{reg}$:
\begin{equation}
E(\mathcal{S}) = E_\text{align}(\mathcal{S}) + w_\text{reg}E_\text{reg}(\mathcal{S}) \text{.}
\end{equation}	
Here, we parameterize the unknown per-pixel scene flow based on a vector $\mathcal{S}$ that stacks the $3G$ unknown stereo $\big\{\mathbf{g}^{s}_k\big\}_{k=1}^G$, motion $\big\{\mathbf{g}_k^m\big\}_{k=1}^G$ and difference flow $\big\{\mathbf{g}_k^d\big\}_{k=1}^G$ control points.
The regularization weight $w_\text{reg}$ balances alignment accuracy with robustness against outliers due to noise and featureless regions.
In the following, we provide details on the employed constraints.

\inlineheading{Spatiotemporal Scene Flow Alignment}
The alignment objective $E_\text{align}$ enforces that the two stereo pairs captured for two subsequent time steps align well in the halfway-domain reference frame.
Since we consider both stereo and temporal motion constraints, this leads to the following spatiotemporal scene flow alignment constraint:
\begin{equation}
E_\text{align}(\mathcal{S}) = w_\text{photo}E_\text{photo}(\mathcal{S}) + w_\text{grad}E_\text{grad}(\mathcal{S}) \text{.}
\label{eq:alignment}
\end{equation}
The quality of alignment of the warped observations in the input frame is quantified based on two terms that model photometric $E_\text{photo}$ (\cref{sec:photometric_alignment}) and gradient-domain $E_\text{grad}$ (\cref{sec:gradient_alignment}) alignment constraints, respectively.
The weights $w_\text{photo}$ and $w_\text{grad}$ define the relative importance of these terms.

\inlineheading{Spatial Regularization Constraints}
Recovering the unknown scene flow $\mathcal{S}$ from the two captured stereo pairs is a challenging problem due to noise in the image acquisition process and ambiguities due to featureless image regions.
To allow for the robust estimation of high-quality scene flow despite these challenges, we propose an efficient regularization strategy based on three terms:
\begin{equation}
E_\text{reg}(\mathcal{S}) = w_\text{smooth}E_\text{smooth}(\mathcal{S}) + w_\text{epi}E_\text{epi}(\mathcal{S}) + w_\text{mag}E_\text{mag}(\mathcal{S}) \text{.}
\label{eq:regularization}
\end{equation}
The first term $E_\text{smooth}$ (\cref{sec:flow_smoothness}) enforces the local smoothness of the estimated flow fields.
This term allows to handle noisy input data and bridges the uncertainty created by missing or incorrect alignment constraints in featureless regions.
The second term $E_\text{epi}$ constrains the stereo flow to be consistent with the epipolar geometry of the binocular camera setup.
Different to previous methods, such as Valgaerts et al. \cite{ValgaBZWST2010}, we do not manually linearize this constraint, leading to a better approximation of the derivatives.
Finally, the third term $E_\text{mag}$ (\cref{sec:flow_magnitude}) constrains the flows to a reasonable magnitude leading to higher robustness.
The weights $w_\text{smooth}$, $w_\text{epi}$ and $w_\text{mag}$ influence the relative importance of the terms.

\subsection{Photometric Alignment}
\label{sec:photometric_alignment}

We enforce the photometric alignment of the captured input images to the halfway-domain reference frame based on a brightness constancy constraint:
\begin{equation}
E_\text{photo}(\mathcal{S}) = \sum_{i=1}^N{\sum_{k=0}^{5}{V(\mathbf{x}_i ) \cdot W(\mathbf{x}_i) \cdot \Phi\Big( d_{k}(\mathbf{x}_i) \Big) }} \text{.}
\label{eq:photometric_alignment}
\end{equation}
%
The \NEW{visibility map} $V(\mathbf{x}_i )$ \NEW{encodes} the visibility of the associated 3D point (1: visible, 0: not visible).
Different from many related methods, we take visibility into account based on computed per-pixel occlusion maps, which are bootstrapped based on a hierarchical optimization scheme (\cref{sec:hierarchy_optimization}).
The weight $W(\mathbf{x}_i)$ is used for pruning outliers based on color similarity: it is one if the residual per-pixel color distance in the reference frame is smaller than a threshold $\epsilon_\text{p} \!=\! 0.2$, and zero otherwise.
The functions $d_k$ are the flow consistency constraints in \cref{sec:consistency}.
Instead of a least-squares formulation, we use the robust \NEW{pseudo-}Huber penalty function for increased robustness against outlier correspondences:
\begin{equation}
\Phi(x) = \sqrt{x^2 + \epsilon^2} \text{.}
\end{equation}
We use $\epsilon \!=\! \text{0.001}$ in all our experiments.

\subsection{Gradient Domain Alignment}
\label{sec:gradient_alignment}

In addition to the photometric alignment term, we also use a gradient domain alignment constraint in the reference frame:
\begin{equation}	
E_\text{grad}(\mathcal{S}) \!=\! \sum_{i=1}^N{\sum_{k=0}^{5}{ V(\mathbf{x}_i ) \!\cdot\! W(\mathbf{x}_i) \!\cdot\! \Phi\Big( \!\norm{ \nabla d_{k}(\mathbf{x}_i) }^2\! \Big)  }} \text{.}
\label{eq:gradient_alignment}
\end{equation}
This gradient domain measure is more robust to differences in the response functions of the used cameras as well as temporal illumination changes than just a brightness constancy term would be.
$\Phi$ again denotes the robust \NEW{pseudo-}Huber penalty function, $V$ encodes visibility and $W$ prunes outliers based on color dissimilarity.

\subsection{Flow Field Smoothness}
\label{sec:flow_smoothness}

To increase the robustness against noise and featureless regions in the input images, we incorporate local smoothness of the three flow fields (motion, stereo and difference flow) by enforcing neighboring displacements to be similar:
\begin{equation}
E_\text{smooth}(\mathcal{S}) = \sum_{i=1}^G{\sum_{j \in \mathcal{N}_i} \sum_{f \in \{\text{s}, \text{m}, \text{d}\}} w_i w_f \norm{ \mathbf{g}^f_i - \mathbf{g}^f_j }^2} \text{.}
\label{eq:flow_smoothness}
\end{equation}
The three weights $w_f$ (for $f \!\in\! \{\text{s}, \text{m}, \text{d}\}$) balance the smoothness of the stereo, motion and difference flow, respectively.
The per-pixel weight $w_i$ takes into account how discriminative a small 3$\times$3 pixel region around the $i^\text{th}$ grid point is and hence how well it can be tracked.
For featureless regions, $w_i$ is set to a high value, which strengthens regularization.
We compute this weight by analyzing \NEW{the two eigenvalues} of the auto-correlation matrix of the image patches.

\subsection{Epipolar Geometry Consistency Constraint}
\label{sec:epi_polar}

For increased robustness and to further constrain the flow fields, we enforce the stereo flow to be consistent with the epipolar geometry of the fixed stereo camera setup:
\NEW{
\begin{equation}
E_\text{epi}(\mathcal{S}) = \sum_{i=1}^G \big[(\mathbf{l}_i^0)^{\!\top} \mathbf{F}(\mathbf{r}_i^0)\big]^2 + \big[(\mathbf{l}_i^1)^{\!\top} \mathbf{F}(\mathbf{r}_i^1)\big]^2 \text{.}
\label{eq:flow_magnitude}
\end{equation}
Here, $\mathbf{F}$ is the fundamental matrix and the 3D vectors $\mathbf{l}_i^t$ and $\mathbf{r}_i^t$ denote the homogeneous coordinates of reference grid point positions $\mathbf{g}^x_i$ transformed to the left and right camera, respectively.
The transformed 2D reference positions are:}
\begingroup\makeatletter\def\f@size{8}\check@mathfonts
\begin{align}
\hat{\mathbf{l}}_i^0 = \mathbf{g}^x_i - \mathbf{g}^s_i - \mathbf{g}^m_i+\mathbf{g}^d_i\text{,} \quad \hat{\mathbf{r}}_i^0 = \mathbf{g}^x_i+\mathbf{g}^s_i- \mathbf{g}^m_i- \mathbf{g}^d_i \text{,}\\
\hat{\mathbf{l}}_i^1 = \mathbf{g}^x_i- \mathbf{g}^s_i+ \mathbf{g}^m_i-\mathbf{g}^d_i\text{,} \quad \hat{\mathbf{r}}_i^1  = \mathbf{g}^x_i+\mathbf{g}^s_i+\mathbf{g}^m_i+\mathbf{g}^d_i \text{.}
\end{align}
\endgroup
This constraint effectively enforces that corresponding pixels in the images are close to the corresponding epipolar line.
Different to previous methods, e.g. Valgaerts et al. \cite{ValgaBZWST2010}, we do not manually linearize this constraint and thus obtain a better approximation of the derivatives.

\subsection{Flow Field Magnitude}
\label{sec:flow_magnitude}

We further stabilize the scene flow estimation by constraining the magnitude of the three flow fields.
This Tikhonov regularization strategy is enforced based on the following soft-constraint:
\begin{equation}
E_\text{mag}(\mathcal{S}) = \sum_{i=1}^G \sum_{f \in \{\text{s}, \text{m}, \text{d}\}} m_f \norm{ \mathbf{g}^f_i }^2 \text{.}
\label{eq:flow_magnitude}
\end{equation}
Since the stereo, motion and difference flows exhibit different properties, we use the weights $m_f$, $f \!\in\! \{s,m,d\}$, to balance these constraints. 
Due to temporal coherence, we assume the motion flow to be smaller than the stereo flow.
The difference flow is assumed to be the smallest, since it only models the residual displacement.
We introduce a hierarchical optimization strategy in \cref{sec:hierarchy_optimization} that still allows to handle large displacements on the coarser levels of the hierarchy.

\subsection{Scene Flow Parameters}
\label{sec:parameters}

The choice of parameters influences our scene flow energy and the reconstruction results.
%
%
Our approach proved quite robust to variation in the specific parameter values.
Nevertheless, the best reconstruction results are obtained at the sweet spot between the data fitting term and the prior constraints.
We provide the parameters used to generate the results in Appendix \ref{sec:appendix}.

%% file: 6optimization.tex

\section{Data-Parallel Optimization}
\label{sec:optimization}

The number of unknowns of our non-linear scene flow objective $E(\mathcal{S}) \colon \mathbb{R}^{6G} \!\to\! \mathbb{R}$ depends on the number of control points $G$ of the used deformation grid (two unknowns for each \NEW{node} of the three different flows).
Since this number directly depends on the image resolution and the grid step size (aka the coarsening factor), this leads to a large number of unknowns, even for smaller image resolutions,
for example $(2 \!\times\! 3 \!\times\! 800 \!\times\! 600) / 2^2 \!=\! \text{720K}$ unknowns for an image resolution of $800 \!\times\! 600$\,pixels and a grid step size of 2\,pixels.
Since we aim to solve the scene flow problem at real-time frame rates, we devise a data-parallel hierarchical solver, following Zollhöfer et al. \cite{ZollhNIRZFWFLTS2014}, that exploits the computational power of modern graphics cards.
Our hierarchy encodes flows based on deltas to the next coarser level.
This enables \NEW{us} to handle large displacements and allows for fast convergence based on a temporal propagation strategy.

We cast finding the scene flow $\mathcal{S}^*$ that best explains the input observations as a non-linear optimization problem:
\begin{equation} \label{eq:opt}
\mathcal{S}^* = \argmin_{\mathcal{S}}{E(\mathcal{S})} \text{.}
\end{equation}
This is a general unconstrained optimization problem, since the alignment objective $E_\text{align}$ (\cref{eq:alignment}) does not fit the canonical least-squares structure of the other objectives due to the robust \NEW{pseudo-}Huber penalty.
Since it is challenging to devise real-time data-parallel solvers for such problems, we transform our problem to a non-linear least-squares problem by taking the square root of the residuals ($x \equiv (\sqrt{x})^2$).

\subsection{Data-Parallel Gauss-Newton Solver}
\label{sec:solver}

After this transformation, the optimization problem fulfills the canonical least-squares form, and can be written as a sum of squared residual terms $r_m$:
\begin{equation}
E(\mathcal{S}) = \sum_{m=1}^{M}{r_m^2(\mathcal{S})}.
\end{equation} 
%
We stack all $M$ residuals into the residual vector operator $\mathbf{R} \colon \mathbb{R}^{6G} \!\to\! \mathbb{R}^M$ and rewrite the energy $E$ using it:
\begin{align}
E(\mathcal{S}) &= \norm{ \mathbf{R}(\mathcal{S}) }^2 \text{,} \\
\mathbf{R}(\mathcal{S}) &= \begin{bmatrix} r_1(\mathbf{\mathcal{S}}) & \hdots & r_{M}(\mathcal{S}) \end{bmatrix}^{\!\top} \text{.}
\end{align}
Our proposed objective function comprises $M\!= 2N + 14G$ residuals $r_m$ due to the used photometric alignment ($N$), gradient-domain alignment ($N$), smoothness ($6G$), epipolar ($2 G$) and flow magnitude ($6G$) constraints.
Due to the large number of residuals ($M$) and unknowns ($6G$), a data-parallel optimization strategy is of paramount importance to achieve real-time frame rates.
Since the residual vector $\mathbf{R}$ is still non-linear in the unknowns $\mathcal{S}$, Gauss-Newton explicitly linearizes $\mathbf{R}$ based on a first-order Taylor expansion:
\begin{equation}
\mathbf{R}(\mathcal{S}_{k+1}) \approx \mathbf{R}(\mathcal{S}_{k})+\mathbf{J}(\mathcal{S}_{k})\cdot \delta,~~~\delta = \mathcal{S}_{k+1}-\mathcal{S}_{k} \text{.}
\label{eq:delta_update}
\end{equation}
Here, $\mathbf{J}(\mathcal{S}_{k})$ is the Jacobian of $\mathbf{R}$ evaluated at the solution after $k$ iterations.
\NEW{The Jacobian is computed based on analytical derivatives.}
The resulting least-squares problem to find optimal updates $\delta^\ast$ is:
\begin{equation}
\delta^\ast=\argmin_\delta{\norm{\mathbf{R}(\mathcal{S}_{k}) + \mathbf{J}(\mathcal{S}_{k})\cdot \delta}^2} \text{.}
\end{equation}
The optimum is computed by solving the associated normal equations based on a data-parallel preconditioned conjugate gradient (PCG) \cite{ZollhNIRZFWFLTS2014} solver.
Similar to Zollhöfer et al. \cite{ZollhDIWSTN2015} \NEW{and Wu et al.} \cite{WuZNSIT2014}, we also employ a domain decomposition strategy for higher performance, and a hierarchical optimization strategy to speed up convergence.
However, in contrast, we employ a hierarchy of delta updates that allows for a better temporal initialization strategy and the computation of large displacements.
This strategy also seamlessly integrates with our flow magnitude constraints $E_\text{mag}$ (\cref{sec:flow_magnitude}), enabling the computation of large flow displacements, since we only encourage the deltas to be small.

\subsection{Domain Decomposition}

We divide the problem into small subproblems based on a subdivision of the halfway-domain reference frame into small square subdomains \NEW{of size 16$\times$16\,pixels (plus a boundary of 2\,pixels)}.
The optimization is then performed using multiple data-parallel \textit{Alternating Schwarz} \citep{Zhao1996,ZollhDIWSTN2015} iterations.
In each iteration, subproblems are locally solved based on one step of data-parallel Gauss-Newton (PCG for linear system), and the subdomain data exchange is handled via global memory.
During PCG, all required data is kept in shared memory for increased performance.

In contrast to previous work \cite{ZollhDIWSTN2015,ZollhNIRZFWFLTS2014,WuZNSIT2014}, we precompute the non-zero entries of $\mathbf{J}^\top\mathbf{J}$ for the alignment term ($9 \!\times\! 3 \!\times\! 2 \!\times\! 2$ per warp grid point) and read them on demand.
This strategy is more efficient than evaluating them on the fly, since the computation of the system matrix for the alignment term is expensive due to the combinatorial explosion caused by every grid point depending on multiple pixels.
Regularizers are still applied on-the-fly in each iteration step.

\subsection{Delta Hierarchy for Fast Optimization}
\label{sec:hierarchy_optimization}

Our optimization strategy works in a coarse-to-fine manner, but in contrast to previous work \cite{ZollhDIWSTN2015,ZollhNIRZFWFLTS2014,WuZNSIT2014}, we use a hierarchy of delta flows (\NEW{still} with a downsampling factor of 2).
\NEW{This means that each level only stores and computes an offset with respect to the next coarser one}.
This helps fast convergence and the computation of large displacement flow fields.
We flip-flop between solving and upsampling the results to the next finer level based on bilinear interpolation until the finest resolution level is reached.
The number of levels used in our hierarchy depends on the resolution of the input images.

In the first frame, all flows are initialized to zero.
In subsequent frames, we initialize the flow fields based on the results obtained in the previous time step.
Based on the assumption of constant velocity, we use the computed motion flow to propagate all flow estimates from the previous to the next time step.
Since we employ a delta hierarchy, we transfer the delta flows on each level separately.

We also use the hierarchy to bootstrap occlusion maps for visibility computation.
To this end, we render the currently estimated geometry on every level from the camera views, and determine all visible pixels based on a \textit{z}-buffer.
The occlusion maps are interpolated to the next finer level and used to prune invisible pixels in the alignment term (\cref{eq:alignment}).

In a similar fashion, the illumination correction is applied on every hierarchy level.
To this end, we compute the intensity residual between the two stereo pairs in the reference frame, and extract the low-frequency components by convolution with a Gaussian filter ($\sigma \!=$\NEW{3.2}\,pixels).
The extracted low-frequency components are attributed to illumination and/or differences in the cameras' response functions.
We upsample the illumination differences using a box filter to the next finer level and use them to normalize the input images.

%% file: 7results.tex

\section{Results}
\label{sec:results}

We evaluate our approach on live data captured using a custom stereo webcam rig and also on publicly available datasets.
The reconstructions based on our stereo rig are obtained at real-time frame rates.
In addition, we apply our approach to the high-resolution, high-quality stereo data of Schneider et al. \cite{SchneKHE2011}, and Blumenthal-Barby and Eisert \cite{BlumenthalBarby201489}.
Our approach scales well to this high-resolution data in terms of reconstruction quality and runtime performance.
We also compare our approach to the slow, but high-quality, off-line scene flow approach of Valgaerts et al. \cite{ValgaBZWST2010}.
Our approach obtains similar quality at much higher frame rate.
%

\subsection{Live Results}

\begin{figure}
	\centering
	\includegraphics[width=\linewidth]{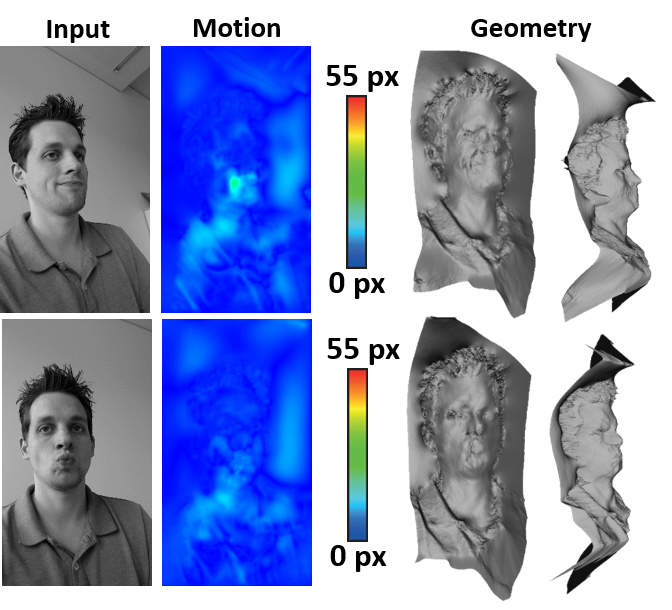}\vspace{-0.2em}
	\caption{
		Live reconstruction results.
	}\vspace{-0.5em}
	\label{fig:live}
\end{figure}

We use two Logitech HD Pro C920 webcams to capture a stereo video stream at 1280$\times$720\,pixels (0.9\,MP).
The cameras' refresh rate is 30\,Hz.
Using our data-parallel solution strategy, we compute the scene flow at the refresh rate of the cameras.
\Cref{fig:live} shows stereo reconstruction results and the corresponding scene flow obtained using our custom stereo webcam rig.
As can be seen, we handle fully dynamic scenes and obtain detailed reconstructions.

\subsection{Runtime Performance and Convergence}
\label{sec:convergence}

\begin{figure}
	\includegraphics[width=\linewidth]{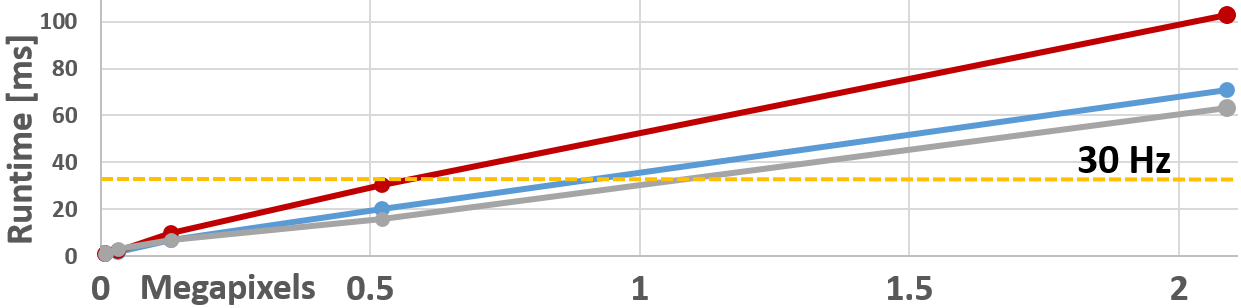}
	\vspace{-0.4cm}
	\caption{\label{fig:timings}
		Runtime performance of our approach with different grid step sizes: red = 1, blue = 2, gray = 4.
	}\vspace{-0.5em}
\end{figure}

\Cref{fig:timings} plots the runtime of our approach with respect to the resolution of the input images and different grid step sizes.
As can been seen, the runtime of our approach scales linearly with the input image resolution.
We obtain full real-time frame rates for up to 0.9\,MP.
For the resolution of our live setup (1280$\times$720\,pixels = 0.9\,MP), we require 31\,ms to compute the scene flow.
This high performance is a direct result of our data-parallel optimization strategy and our specifically tailored scene flow objective.
For all timings, we used 5 hierarchy levels.
On the two finest levels, we perform 2 non-linear iterations.
On all other levels, we perform 5 non-linear iteration steps.
In each non-linear Gauss-Newton step, we use 5 PCG iterations (with 5 patch iterations each) to solve the underlying system of normal equations.

\begin{figure}
	\includegraphics[width=\linewidth]{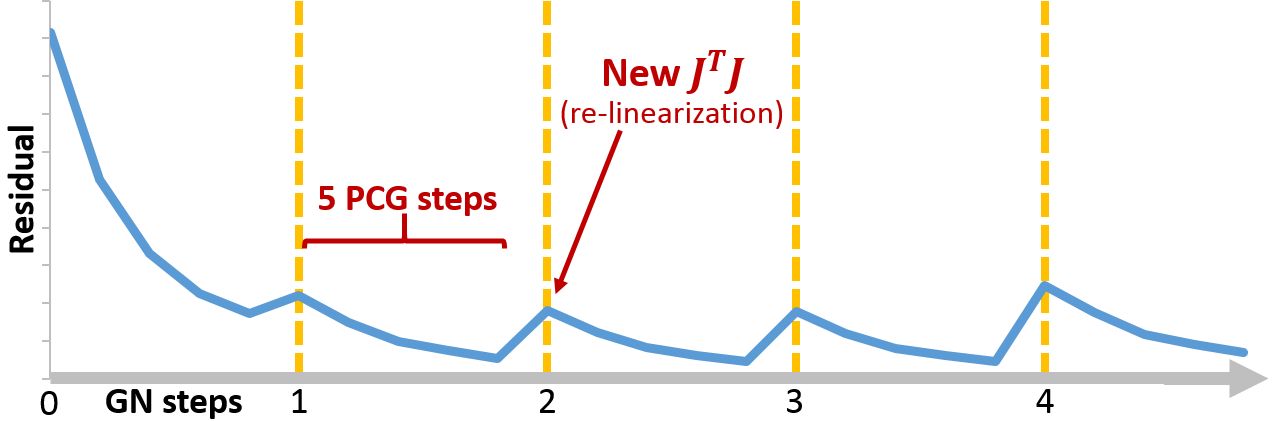}
	\vspace{-1em}
	\caption{\label{fig:conv}%
		Convergence plot.
		See \protect\cref{sec:convergence} for discussion.
	}\vspace{-0.5em}
\end{figure}

We next analyze the convergence behavior of our solver for the finest hierarchy level.
To this end, we reconstructed the scene flow between two time steps.
In this evaluation, we apply 5 non-linear Gauss-Newton steps with 5 PCG steps (each using 5 patch iterations).
\Cref{fig:conv} plots the linear residual of the normal equations.
For each single non-linear step, the error is always decreased by the PCG iteration steps.
The error peaks every 5 steps, which marks the beginning of each new non-linear Gauss-Newton iteration.
At these points, the problem is newly linearized using Taylor series expansion, leading to new normal equations.
Therefore, the error of this new system is higher, but it is directly decreased in the following iterations.
Note that these new systems are better approximations of the real function, since the linearization is performed closer to the optimum.

\subsection{Comparison to Valgaerts et al. \cite{ValgaBZWST2010}}

\begin{figure}
	\includegraphics[width=\linewidth]{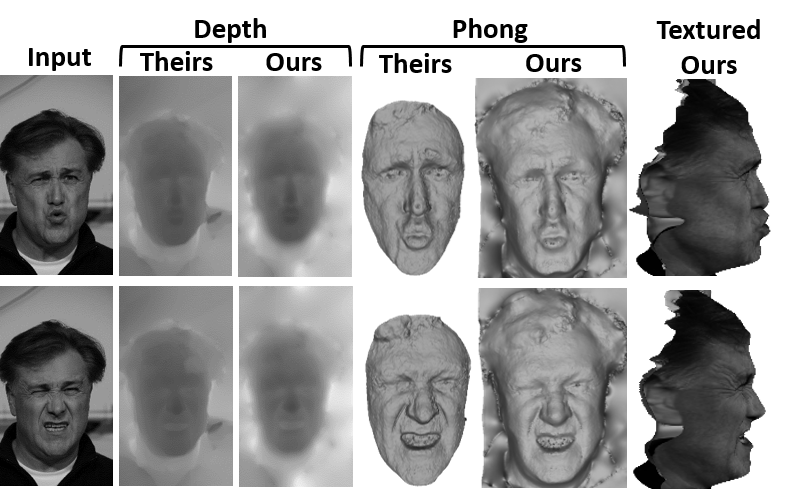}
	\vspace{-1em}
	\caption{\label{fig:levi}
		Comparison to the approach of Valgaerts et al. \cite{ValgaBZWST2010} \NEW{on the} `Volker' dataset \cite{ValgaWBST2012}.
	}\vspace{-0.5em}
\end{figure}

\begin{figure*}
	\includegraphics[width=\linewidth]{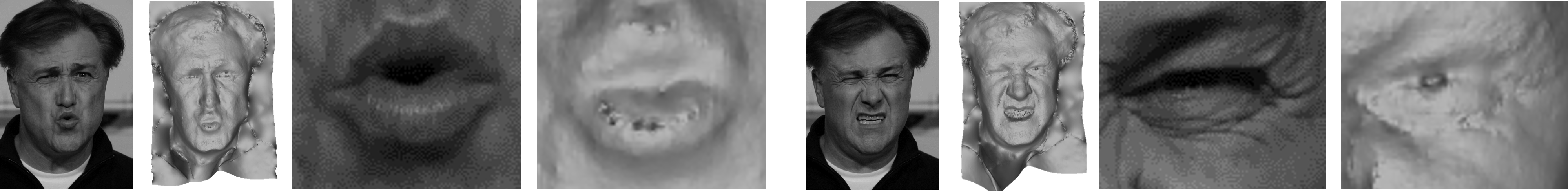}
	\vspace{-1em}
	\caption{\label{fig:volker}
		Our high-quality reconstruction results \NEW{on the} `Volker' dataset \cite{ValgaWBST2012}.
	}\vspace{-0.5em}
\end{figure*}

\begin{figure}
	\includegraphics[width=\linewidth]{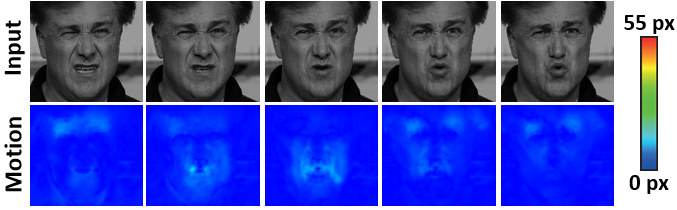}
	\vspace{-1em}
	\caption{\label{fig:motion_flow_volker}
		Our motion flow on the `Volker' dataset \cite{ValgaWBST2012}.
	}
\end{figure}

We compare our approach to the slow, but high-quality off-line state-of-the-art scene flow approach of Valgaerts et al. \cite{ValgaBZWST2010}, see \cref{fig:levi}.
As can be seen, our approach obtains \NEW{reconstructions of similar quality}.
Note that our approach is three orders of magnitudes faster than theirs.
Due to the high resolution of the input data (1920$\times$1088\,pixels = 2.1\,MP), we use 5 non-linear iterations with 5 PCG steps (each with 5 patch iterations).
The employed hierarchy uses 5 levels, as before.
With our approach, we obtain the scene flow for a single pair of frames in only 110\,ms, while Valgaerts et al.'s method \cite{ValgaBZWST2010} requires more than 6 minutes per frame (more than 3,000 times slower).
%
We attribute this performance advantage of our approach to the smart design of our objective function, including the reduction of unknowns through our coarser warp grid, that allows to apply our highly efficient data-parallel non-linear least-squares framework.
The geometry and motion obtained by our approach is of very high quality, as shown in \cref{fig:volker} and \cref{fig:motion_flow_volker}.

\begin{figure*}
	\includegraphics[width=\linewidth]{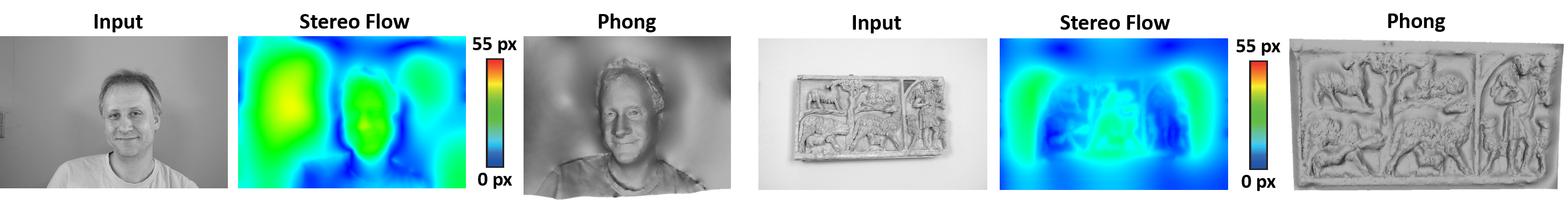}
	\vspace{-1.5em}
	\caption{\label{fig:stereo}
		Our high-quality reconstruction results on the data of Schneider et al.~\protect\cite{SchneKHE2011} (left) and Blumenthal-Barby and Eisert \cite{BlumenthalBarby201489} (right).}
\end{figure*}

\begin{figure}
	\vspace{-1em}
	\includegraphics[width=\linewidth]{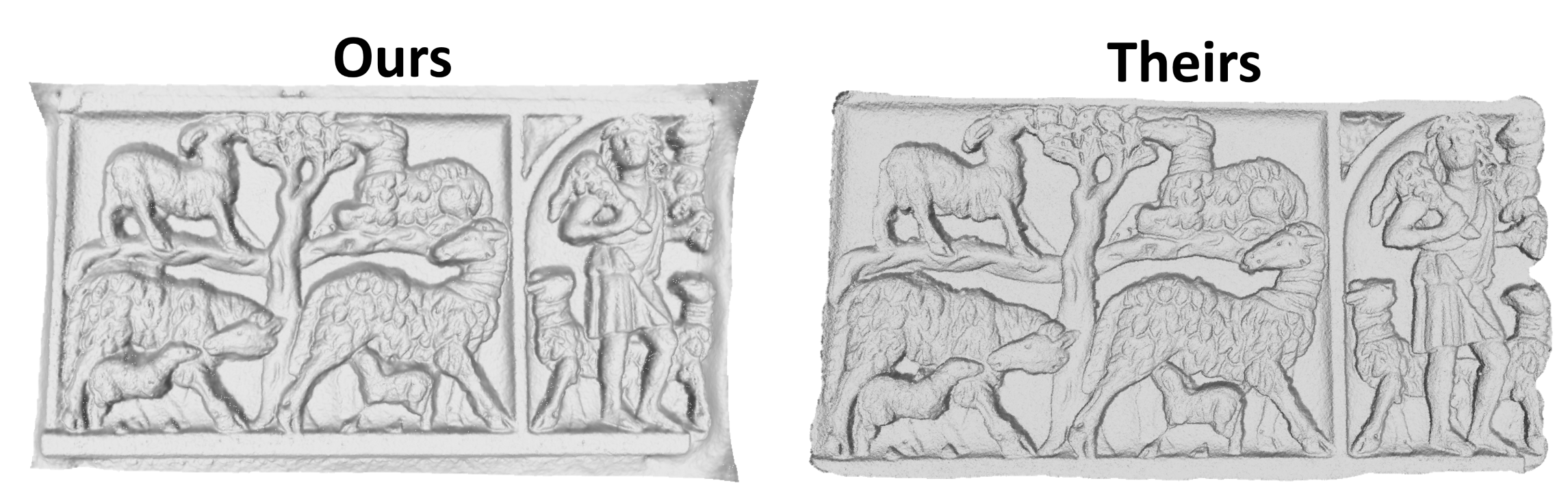}
	\vspace{-1em}
	\caption{\label{fig:comp_stereo}
		Comparison to Blumenthal-Barby and Eisert \cite{BlumenthalBarby201489}.
	}
\end{figure}

\subsection{Comparison to Warping-based Approaches}

We also applied our approach to the stereo reconstruction problem.
To this end, we use publicly available high-quality stereo images \cite{SchneKHE2011,BlumenthalBarby201489} with an input image resolution of 4288$\times$2848\,pixels (12\,MP).
Since only a single stereo pair is available for these scenes, we initialize both time steps using the same image pair.
Our approach obtains high-quality reconstruction results, shown in \cref{fig:stereo}, which are on par with previous \NEW{approaches}, see also \cref{fig:comp_stereo}.
We obtain these results at a much shorter computation time.
Our approach requires only 700\,ms for reconstruction, and is two orders of magnitude faster than the approach of Blumenthal-Barby and Eisert \cite{BlumenthalBarby201489}, which requires several minutes.
We attribute this difference in runtime performance to the design of our energy function and our data-parallel solution strategy.
Note that if a sequence of stereo frames is available, our approach is also able to estimate the motion flow.

%
%



%

%% file: 8conclusion.tex

\section{Limitations}
\label{sec:limitations}

Our approach obtains high-quality scene flow at real-time frame rates.
Nevertheless, it is subject to a few limitations.
We summarize them here and give ideas for future work in this domain.
Sometimes our mip-map-based hierarchical downsampling strategy is too coarse.
Therefore, distinctive regions are lost on coarser levels.
This complicates the scene flow computation.
A finer hierarchy in combination with feature-preserving downsampling \cite{BoussPD2009} could alleviate this problem.
Currently, our stereo setup has to be precalibrated before use.
This is a cumbersome process and has to be repeated every time the camera setup changes.
In the future, methods could be investigated to jointly optimize for the extrinsic camera parameters to allow for fully dynamic camera setups\NEW{, similar to Valgaerts et al. \cite{ValgaBZWST2010}}.
Like every other passive stereo reconstruction approach, our approach suffers from problems in featureless regions of the scene.
In these regions, the data term is not sufficiently discriminative and the used regularization terms take over.
If the scene violates these prior assumptions, the obtained reconstructions do not match reality.
Currently, in the first frame, we initialize the stereo and motion flow to zero. Therefore, our approach sometimes needs a few frames to converge.
In the future, smart initialization strategies could be explored to jump-start the optimization process from the very first frame.

\section{Conclusion}
\label{sec:conclusion}

We presented an approach for real-time joint reconstruction of motion and geometry from stereo RGB videos.
To this end, we extended the concept of the halfway domain to scene flow.
Our approach achieves real-time performance based on a novel data-parallel solver that exploits the computational horsepower of modern graphics cards.
Comparisons and evaluations show that high-quality scene flow estimates can be obtained at the cameras' refresh rate using variational optimization.
We believe that the availability of scene flow data at real-time frame rates is an important building block for many other approaches, such as real-time non-rigid structure-from-motion.